%% file: main.tex
\documentclass[runningheads]{llncs}

\usepackage{latex_src/eccv}

\usepackage{siunitx}
\usepackage{amsmath,amssymb,amsfonts}
\usepackage{mathtools}
\usepackage{dsfont}
\usepackage{bm}
\usepackage{xcolor}
\usepackage{enumitem}
\definecolor{GammaColor}{RGB}{180,0,200} 
\newcommand{\method}{MVDG}

\usepackage{latex_src/eccvabbrv}

\usepackage{graphicx}
\usepackage{booktabs}
\usepackage{wrapfig}
\usepackage{float}
\usepackage{placeins}

\usepackage[accsupp]{axessibility}  

\usepackage{hyperref}

\usepackage{orcidlink}

\begin{document}

\title{\method: Efficient Multi-view 3D Disambiguation on Unconstrained Real-World Images} 


\author{Hanyuan Xiao\inst{1,2} \and
Gonglin Chen\inst{1,2} \and
Haolin Xiong\inst{1,2} \and
Wenbin Teng\inst{1,2} \and \\
Haiwei Chen\inst{1,2} \and
Yajie Zhao\inst{1,2}
}

\authorrunning{H.Xiao et al.}
\titlerunning{MVDG}

\institute{University of Southern California \and
Institute for Creative Technologies
}

\maketitle

\input{figures/tex/teaser}

\input{sec/0_abstract}
\input{sec/1_introduction}
\input{sec/2_related_works}
\input{sec/3_preliminaries}

\input{sec/4_method}
\input{sec/5_experiments}
\input{sec/6_discussion}

\FloatBarrier

%
%
\bibliographystyle{latex_src/splncs04}
\bibliography{references}

\input{sec/X_suppl}

\end{document}

%% file: figures/tex/teaser.tex
\vspace{-1cm}
\begin{figure}
      \centering
      \includegraphics[width=1.0\linewidth]{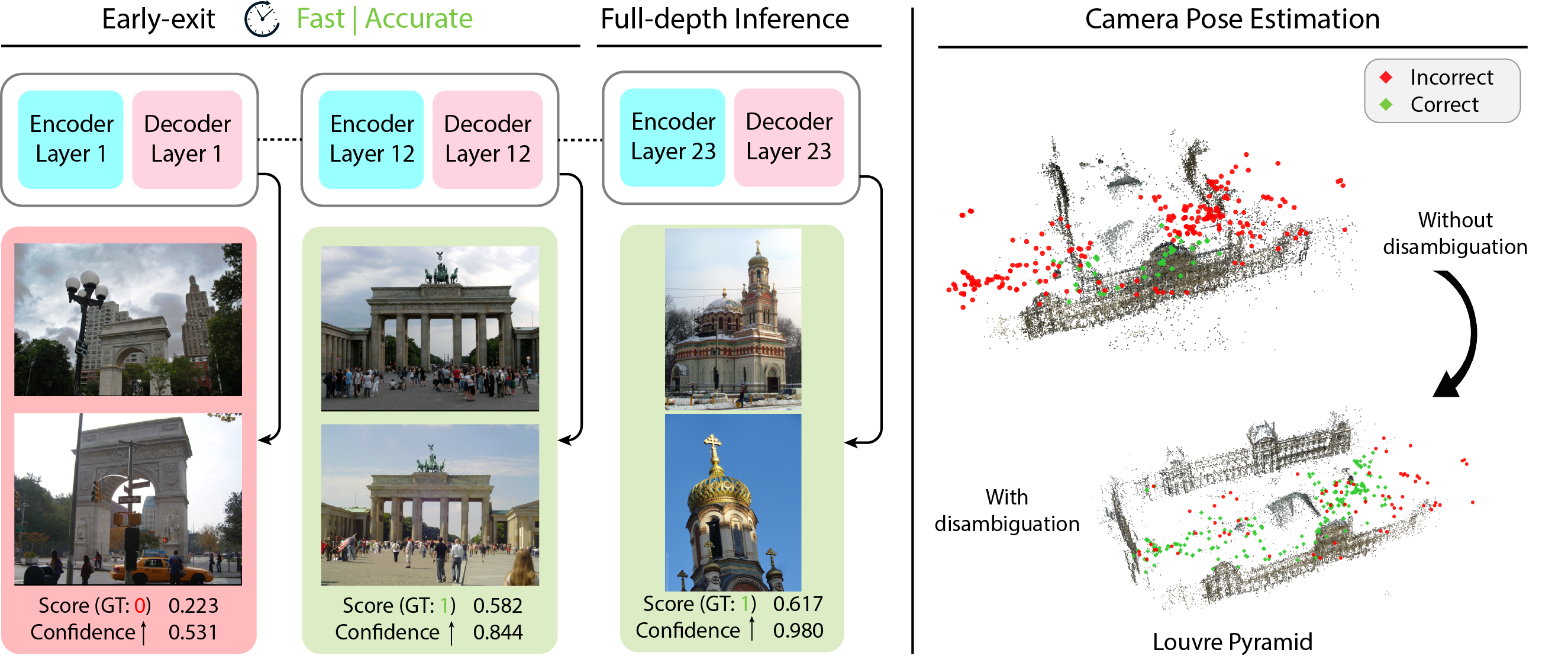}
      \caption{\textbf{Overview of \method\ for multi-view 3D disambiguation.} \method classifies image pairs with similar 2D appearance that may correspond to different viewpoints. We train a classification decoder and a confidence module to select effective encoder depths adaptively. By exiting at early layers when confidence is high, our method improves efficiency while maintaining accuracy. (\textit{Left}) \textcolor{red}{Red}-boxed examples are false pairs (different scene/view), and \textcolor{green}{green}-boxed examples are true pairs (same scene/view). (\textit{Right}) We show a comparison of structure-from-motion result without vs. with \method\ to remove visually ambiguous images.}
      \label{fig:teaser}
\end{figure}
\vspace{-1cm}


%% file: sec/0_abstract.tex
\begin{abstract}

Illusory matches between distinct yet visually similar 3D surfaces---\emph{doppelgangers}---remain a fundamental obstacle for large-scale, in-the-wild 3D reconstruction and visual localization. Prior work mitigates this issue with pairwise classifiers, but this design limits multi-view contextual reasoning and incurs $\mathcal{O}(n^2)$ inference complexity for downstream structure-from-motion (SfM). We present \method, a scalable multi-view disambiguation framework built on the 3D foundation model VGGT, which jointly reasons over an arbitrary number of multiview images. By incorporating 3D-aware multi-view features, our method reduces dependence on pairwise comparisons by encoding and decoding views in a single pass. We further observe that direct multi-view fine-tuning of VGGT can be unstable under noisy supervision; motivated by label ambiguity in \emph{Doppelgangers}, we construct a pseudo-pairwise training set from \emph{AerialMegaDepth} and show that fine-tuning on sampled subsets yields stable optimization and strong generalization to held-out scenes. Finally, because full SfM evaluation (even with faster pipelines such as GLOMAP) remains expensive, we process a pseudo-pairwise dataset for efficient validation; we derive a predictive relationship between regular SfM metrics and the classification accuracy on this pseudo-pairwise test. Experiments show that our method achieves comparable pairwise accuracy while improving both SfM accuracy and inference speed over baselines.

\keywords{Doppelgangers \and 3D foundation model \and Structure-from-motion}

\end{abstract}

%% file: sec/1_introduction.tex
\section{Introduction}
\label{sec:introduction}
\input{figures/tex/motivation}

Recovering coherent 3D structure from multi-view imagery is a central goal in computer vision. Conventional pipelines rely on hand-crafted local features and robust 2D correspondence estimation for 3D point reconstruction in Structure-from-Motion (SfM) and Multi-View Stereo (MVS)~\cite{5539802,schoenberger2016sfm,schoenberger2016mvs,10.1145/1179352.1141964}. In contrast, recent data-driven approaches~\cite{wang2025vggtvisualgeometrygrounded,wang2024dust3rgeometric3dvision,wang2025continuous3dperceptionmodel} largely \emph{bypass} explicit 2D feature matching by tokenizing images and decoding geometry in a feed-forward manner, enabling direct inference of 3D properties (e.g., camera parameters, depth, tracking) without iterative optimization. Despite their different paradigms, both families remain vulnerable to the \emph{doppelgangers} problem: when non-covisible viewpoints capture visually similar surfaces, the model may incorrectly match those views and reconstruct inconsistent geometry. This bottleneck has long hindered downstream tasks and real-world deployments.

A natural formulation treats doppelganger detection as binary classification: image pairs with sufficient covisibility (\ie, overlap in visible 3D points) and rigid transformations are labeled true, and pairs below a threshold are labeled false. \cite{cai2023doppelgangerslearningdisambiguateimages} trains a pairwise classifier to reject non-covisible image pairs using differentiable feature evidence; the filtered image sets can then be passed to downstream 3D pipelines such as COLMAP~\cite{schoenberger2016sfm,schoenberger2016mvs}. To leverage pre-trained 3D foundation models, \cite{xiangli2025doppelgangersimprovedvisualdisambiguation} incorporates feed-forward 3D features from~\cite{leroy2024groundingimagematching3d} for pairwise disambiguation. While effective, pairwise designs underuse multi-view constraints and scale poorly in large-scale image collections, where cross-view consistency can expose aliasing contradictions (Fig.~\ref{fig:motivation}). They also incur substantial inference overhead when enumerating all candidate pairs.

We address these limitations with \method, a scalable multi-view 3D disambiguation framework built on VGGT~\cite{wang2025vggtvisualgeometrygrounded}. First, we bridge the data gap between pairwise supervision and multi-view inference. Both \emph{Doppelgangers} and \emph{VisymScenes} are pairwise datasets~\cite{cai2023doppelgangerslearningdisambiguateimages,xiangli2025doppelgangersimprovedvisualdisambiguation}; we convert them into coarse multi-view groups by transferring positive pair labels to anchor-centric neighborhoods. However, annotation noise in \emph{Doppelgangers} introduces ambiguity (Sec.~\ref{sec:limitations}), which makes convergence less stable. We therefore derive a pseudo-pairwise dataset from \emph{AerialMegaDepth}~\cite{vuong2025aerialmegadepth} and show that fine-tuning on these subsets improves training stability and generalization to held-out scenes. Since end-to-end SfM validation at each checkpoint is prohibitively expensive, we also establish an empirical relationship between pseudo-pairwise classification accuracy and SfM metrics (e.g., inlier ratios), yielding a practical proxy for monitoring convergence. Second, to accelerate inference, we train a confidence-based early-exit module that avoids unnecessary deeper-layer computation when early encoder and decoder features already provide reliable predictions. Overall, this design reduces SfM-stage runtime by approximately $40\%$.

In summary, our contributions are:
\begin{enumerate}
    \item We present \method, a scalable multi-view 3D disambiguation framework built on VGGT, and augment pairwise training resources with pseudo-pairwise supervision from \emph{AerialMegaDepth} to improve training stability and generalization.
    \item We establish an empirical relationship between pseudo-pairwise classification accuracy and downstream SfM metrics (e.g., inlier ratio), enabling practical validation without running full SfM. 
    \item We introduce an early-exit inference strategy that halts unnecessary deeper-layer computation when confidence is sufficient, reducing runtime overhead by $\sim 40\%$ on SfM evaluation.
\end{enumerate}

%% file: figures/tex/motivation.tex
\begin{figure}[htb]
      \centering
      \includegraphics[width=\linewidth]{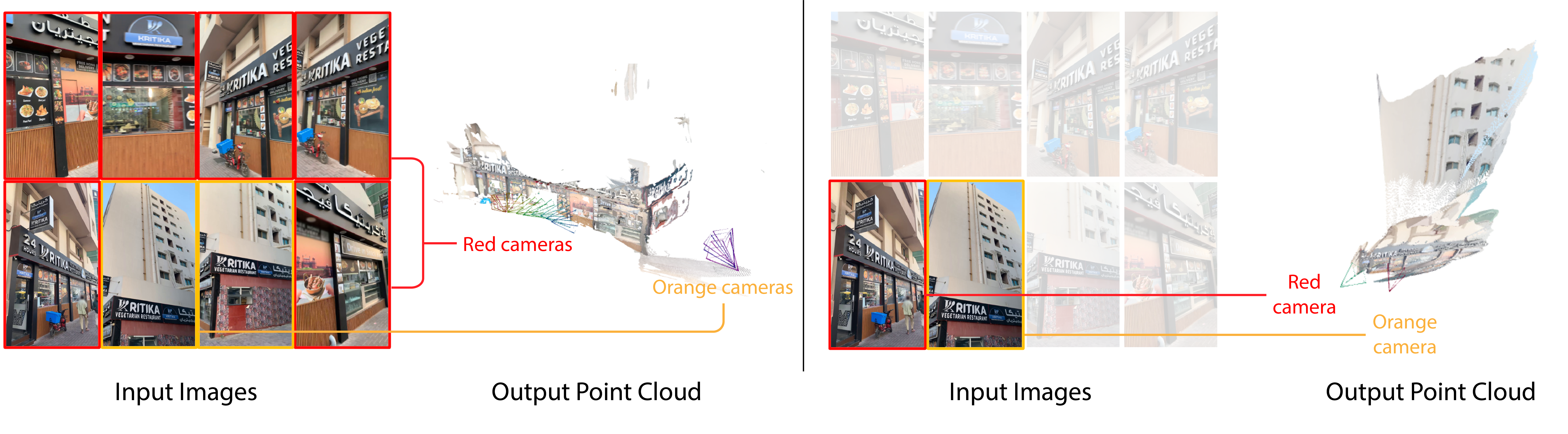}
      \caption{\textbf{Ambiguous inputs can mislead 3D scene understanding.} We feed VGGT~\cite{wang2025vggtvisualgeometrygrounded} with multi-view images from a \emph{VisymScenes} scene~\cite{xiangli2025doppelgangersimprovedvisualdisambiguation} containing repetitive patterns. Among the eight input images, two viewpoint groups are evident to human observers (all from the same physical scene): six \textcolor{red}{red}-marked images and two \textcolor{orange}{orange}-marked images. (\emph{Left}) When all views, including transition geometry, are provided, the backbone robustly disambiguates views and reconstructs a coherent scene. (\emph{Right}) When only two views from different viewpoints are provided, VGGT fails to disambiguate and reconstructs an incorrect joint scene.} 
      \label{fig:motivation}
\end{figure}

%% file: sec/2_related_works.tex
\section{Related Works}
\label{sec:related-works}

\paragraph{Classical SfM/MVS and the limits of 2D matching.}

Early pipelines for image-based 3D reconstruction---typified by COLMAP---combine robust local feature detection and matching (e.g., SIFT) with incremental Structure-from-Motion (SfM) and dense Multi-View Stereo (MVS)~\cite{schoenberger2016sfm,schoenberger2016mvs,journals/ijcv/Lowe04}. COLMAP remains a standard open-source baseline; its core contributions include robust incremental SfM and pixelwise view-selection MVS with strong completeness and accuracy across datasets. These systems scale to large, unordered photo collections and set competitive baselines on multi-view depth and point-cloud benchmarks~\cite{Schops_2017_CVPR}. However, classical pipelines rely on 2D correspondences~\cite{yi2018learninggoodcorrespondences,sarlin2020supergluelearningfeaturematching,edstedt2024roma} and are vulnerable to visual aliasing---repetitive or near-duplicate textures---which can trigger catastrophic misassociations and broken reconstructions. Recent learned matchers improve robustness and efficiency, including SuperPoint~\cite{detone2018superpointselfsupervisedpointdetection}, attention-based SuperGlue~\cite{sarlin2020supergluelearningfeaturematching}, and detector-free LoFTR~\cite{sun2021loftrdetectorfreelocalfeature}. Still, under severe symmetry and repetition, short-range 2D evidence alone remains insufficient.

\paragraph{Learned multi-view geometry and feed-forward 3D backbones.}

Recent feed-forward 3D geometry networks directly infer scene attributes from raw images, moving beyond iterative geometric optimization loops. ~\cite{wang2024dust3rgeometric3dvision} casts pairwise reconstruction as camera-agnostic point-map regression, enabling robust wide-baseline alignment without priors. ~\cite{leroy2024groundingimagematching3d} further grounds matching in 3D by augmenting~\cite{wang2024dust3rgeometric3dvision} with dense local features and a fast reciprocal matcher, improving both accuracy and speed on challenging localization benchmarks. ~\cite{wang20243dreconstructionspatialmemory} introduces a spatial-memory mechanism that predicts per-image point maps in a \emph{global} coordinate frame, eliminating post-hoc global alignment and inferring frame sequences in real time. In parallel, ~\cite{wang2025continuous3dperceptionmodel} models a persistent state with a continuously updated transformer, producing metric point maps online as new views arrive. Most recently, VGGT generalizes this paradigm from one to hundreds of views and directly predicts camera parameters, multi-view depth, dense point maps, and 3D tracks~\cite{wang2025vggtvisualgeometrygrounded}. Despite these gains, computation scales unfavorably with the number of frames because multi-view attention is quadratic in tokens. A concurrent study accelerates VGGT via token-redundancy reduction~\cite{shen2025fastvggttrainingfreeaccelerationvisual}, but it does not explicitly address \emph{doppelgangers} because ambiguous patch mismatches are not modeled. We take a complementary route: confidence-conditioned, depth-adaptive computation for multi-view disambiguation.

\paragraph{Visual aliasing and Doppelgangers.}

~\cite{cai2023doppelgangerslearningdisambiguateimages} formalized the \emph{Doppelgangers} problem and proposed a pairwise classifier trained on a curated dataset, where positive/negative supervision is tied to covisibility and relative rigid-transformation consistency, using keypoints and matches to estimate affine consistency cues.
~\cite{xiangli2025doppelgangersimprovedvisualdisambiguation} improved robustness by leveraging 3D-aware features from~\cite{leroy2024groundingimagematching3d} within a transformer classifier and demonstrated gains when integrating into SfM pipelines across real-world scenes. These works, however, remain pairwise: they do not exploit the richer geometric constraints available when more than two views are jointly encoded, and consequently may under-utilize multi-view redundancy that can disambiguate aliasing. SALAD~\cite{Izquierdo_CVPR_2024_SALAD} serves as a strong 2D retrieval baseline in our comparisons, isolating the gap between global visual similarity and strict 3D covisibility disambiguation. Our approach targets this gap by operating directly on a multi-view 3D backbone, aggregating layer-wise tokens across arbitrary numbers of frames.




%% file: sec/3_preliminaries.tex
\section{Preliminaries}
\label{sec:preliminaries}

Our method builds on VGGT~\cite{wang2025vggtvisualgeometrygrounded}, a 3D foundation model that supports diverse downstream geometry tasks. Given a set of multi-view images, VGGT predicts task-dependent 3D attributes through lightweight decoder heads. The backbone is a transformer encoder with $L=24$ attention blocks, each combining frame attention and global attention. Initialized from DINOv2~\cite{oquab2024dinov2learningrobustvisual} features, the network transitions from local 2D correspondence cues in shallow layers to stronger geometric reasoning in deeper layers. Consequently, tokens aggregated from selected layers (default: 4, 11, 17, 23) provide rich multi-scale information for downstream decoding. However, for disambiguation, it remains unclear which layers best separate visually ambiguous but non-covisible image pairs.

%% file: sec/4_method.tex
\section{Method}
\label{sec:method}

\input{figures/tex/architecture}

\subsection{Overview} \label{sec:method-overview}

Given an unordered image set $\mathcal{I}=\{I_i\}_{i=1}^{N}$ of a static 3D scene, our task is to output covisibility by predicting a binary classification label $s_{i,j}\in\{0,1\}$ for pairs $(I_i, I_j)$. We input all views through a frozen VGGT encoder to obtain layerwise multi-view tokens. After each encoder block, a decoder takes the patch-wise tokens as input and learns to output patch features capable of disambiguation. The decoded features are pooled into a global descriptor vector $\mathbf{f}(I)$ within the scene context for each input image. Pairwise features formed from $(\mathbf{f}(I_i), \mathbf{f}(I_j))$ are classified by an MLP head together with a confidence-driven early-exit module. If the confidence exceeds a threshold, we halt further encoding and take the current prediction as the final decision; otherwise, we continue into later layers. This difficulty-adaptive depth reduces computation for easy cases while still leveraging deeper-layer tokens for challenging ones.

\subsection{Multi-view Supervision from Pairwise Data} \label{sec:method-data}
Existing doppelganger datasets are pairwise~\cite{cai2023doppelgangerslearningdisambiguateimages,xiangli2025doppelgangersimprovedvisualdisambiguation}. The raw input lists pairs for a given anchor image, with between 2 and 88 images paired to the same anchor. We therefore organize the data as a dictionary indexed by image, where each key $I_i$ maps to the set of its paired images $\{I_j\}$ along with their labels $y_{i,j} \in \{0,1\}$ (both positive and negative). This preserves the original pairwise supervision while exposing an anchor-centric multi-view neighborhood during training.

\paragraph{Pseudo-pairwise supervision.}
For large-scale multi-view scenes, we derive pseudo-pairwise labels using geometric overlap computed from available 3D or camera information. We define a covisibility ratio
\begin{equation}
\label{eq:covis-ratio}
\rho_{i,j} = \frac{|\mathcal{P}_i \cap \mathcal{P}_j|}{|\mathcal{P}_i \cup \mathcal{P}_j|},
\end{equation}
and assign
\begin{equation}
\label{eq:pseudo-label}
y_{i,j} = \mathds{1}[\rho_{i,j} \ge \gamma],
\end{equation}
where $\mathcal{P}_i$ denotes the visible 3D keypoints for $I_i$ extracted by ~\cite{schoenberger2016sfm,schoenberger2016mvs} and $\gamma$ is a threshold. We compute $\mathcal{P}_i$ by projecting the ground-truth 3D point cloud provided in \emph{AerialMegaDepth} into the ground-truth camera view of $I_i$. We evaluate discrete thresholds in $\{0.1, 0.2, \ldots, 1.0\}$ and empirically choose $\gamma=0.4$ because it stabilizes multi-view training.
We use \emph{AerialMegaDepth} to fine-tune the model after pairwise training on \emph{Doppelgangers} and \emph{VisymScenes} has converged but SfM accuracy remains unstable. These transferred and pseudo-pairwise labels supervise decoder training in Sec.~\ref{sec:method-module-1}.


\subsection{Multi-view Decoder \& Covisibility Classifier} \label{sec:method-module-1}

Let $L=24$ denote the total number of encoder layers and $\mathbf{T}^{(l)}$ the multi-view tokens at layer $l$. We pass $\mathcal{I}$ through the frozen VGGT encoder to obtain layerwise multi-view tokens $\mathbf{T}^{(l)} = \{T^{(l)}_1, \dots, T^{(l)}_N\}$ for $l=1,\dots,L$. Our goal is to use these intermediate representations to predict whether each pair $(I_i, I_j)$ is covisible. Note that we train the decoder and scoring head through all $L$ layers in this stage.

\paragraph{Per-layer image descriptors.}
For any image $I_i$ and layer $l$, a DPT-style head maps the corresponding tokens to $P$ refined patch descriptors $\{ \mathbf{f}^{(l)}_{p}(I_i) \}_{p=1}^{P}$. We aggregate them into a single image-level vector:
\begin{equation}
\label{eq:img-desc}
\mathbf{f}^{(l)}(I_i) \;=\; \operatorname{LN}\!\left(\frac{1}{P}\sum_{p=1}^{P} \mathbf{f}^{(l)}_{p}(I_i)\right) \in \mathbb{R}^{C},
\end{equation}
where $\operatorname{LN}$ denotes LayerNorm.

\paragraph{Pairwise feature construction.}
To classify the pair $(I_i, I_j)$ at depth $l$, we construct a symmetric, interaction-rich representation that captures agreements and discrepancies:
\begin{equation}
\label{eq:pair-feat}
\begin{aligned}
\mathbf{g}^{(l)}_{i,j}
=
\mathrm{PE}(l) \;+\; \big[
\mathbf{f}^{(l)}(I_i),\;
\mathbf{f}^{(l)}(I_j),\;
\text{abs}(\mathbf{f}^{(l)}(I_i) - \mathbf{f}^{(l)}(I_j)),\;
\mathbf{f}^{(l)}(I_i) \odot \mathbf{f}^{(l)}(I_j)
\big],
\end{aligned}
\end{equation}
where $\odot$ denotes element-wise product and $\mathrm{PE}(l)$ is a positional encoding of the layer index to make the classifier layer-aware.

\paragraph{Per-layer MLP scoring.}
We feed $\mathbf{g}^{(l)}_{i,j}$ into a lightweight MLP classifier to obtain a covisibility probability:
\begin{equation}
\label{eq:mlp-logit}
z^{(l)}_{i,j} = \operatorname{MLP}_{\mathrm{cls}}\!\left(\mathbf{g}^{(l)}_{i,j}\right), \quad
p^{(l)}_{i,j} = \operatorname{Sigmoid}\!\left(z^{(l)}_{i,j}\right).
\end{equation}
A binary covisibility label at depth $l$ is then
\begin{equation}
\label{eq:mlp-thresh}
s^{(l)}_{i,j} = \mathds{1}[p^{(l)}_{i,j} \ge \theta].
\end{equation}
During training we supervise predictions from \emph{all} layers ($l=1,\dots,L$) to maximize attainable accuracy and to enable calibrated early exits later. At inference we select depth adaptively (Sec.~\ref{sec:method-module-2}). We set $\theta=0.5$ for pairwise evaluation and $\theta=0.8$ for SfM evaluation to include positive images more strictly; both values match the baseline Doppelg\"{a}ngers protocol for fair comparison.

\subsection{Confidence-Driven Early-Exit Gating} \label{sec:method-module-2}
We observed that distinctive non-covisible pairs can be rejected using shallow features, while challenging pairs benefit from deeper geometric aggregation (Sec.~\ref{sec:ablation}). Therefore, in the second stage of training, we attach a confidence head that estimates whether the \emph{current} prediction would agree with the \emph{final} (full-depth) prediction.

\paragraph{Confidence head.}
For each layer $l$ and pair $(I_i, I_j)$, we compute
\begin{equation}
\label{eq:conf}
\sigma^{(l)}_{i,j} \;=\; \operatorname{Sigmoid}\!\left(\operatorname{MLP}_{\mathrm{conf}}\!\left(\mathbf{g}^{(l)}_{i,j}\right)\right),
\end{equation}
interpreted as the probability that the decision at depth $l$ matches the decision at depth $L$. At inference we use a fixed threshold $\tau \in (0,1)$: if $\sigma^{(l)}_{i,j} \ge \tau$, we \textbf{exit} at layer $l$ and take $p^{(l)}_{i,j}$ as final; otherwise we proceed to $l{+}1$. We empirically choose $\tau=0.8$ as the confidence threshold considering the marginal gain in prediction accuracy (Sec.~\ref{sec:ablation}).

\subsection{Relationship between pseudo-pairwise accuracy and SfM inliers} \label{sec:method-sfm}

We use pseudo-pairwise labels to measure classification quality on held-out scenes and relate it to downstream SfM outcomes. Let $\mathcal{S}$ denote a set of scenes and $\mathcal{P}_s$ the evaluated image pairs in scene $s$.

\paragraph{Motivation.}
We observe that even after pairwise classification training and validation converge, SfM accuracy (e.g., inlier ratios) can remain unstable and oscillate between high and low values during training. Running full SfM evaluation during validation is infeasible because it is time-consuming (even with GLOMAP) and can exceed GPU memory. We therefore study a correlation between pseudo-pairwise accuracy and SfM inlier ratio as a lightweight proxy: when pseudo-pairwise accuracy is high, downstream SfM accuracy tends to be high as well. This provides a practical signal to visualize and validate convergence on the SfM task without running the full pipeline each time.

\paragraph{Pseudo-pairwise accuracy.}
We compute per-scene accuracy as
\begin{equation}
\label{eq:pp-acc}
\mathrm{Acc}_s = \frac{1}{|\mathcal{P}_s|}\sum_{(i,j)\in\mathcal{P}_s} \mathds{1}\left[\hat{s}_{i,j}=y_{i,j}\right].
\end{equation}

\paragraph{SfM inlier ratio.}
We summarize SfM quality with an inlier ratio
\begin{equation}
\label{eq:sfm-inlier}
\mathrm{Inlier}_s = \frac{|\mathcal{I}^{\mathrm{inlier}}_s|}{|\mathcal{I}^{\mathrm{input}}_s|},
\end{equation}
where $\mathcal{I}^{\mathrm{input}}_s$ denotes the set of input cameras after doppelganger-based filtering, and $\mathcal{I}^{\mathrm{inlier}}_s$ denotes the reconstructed inlier cameras. 

\paragraph{Predictive relationship.}
We fit a saturating exponential mapping from pseudo-pairwise accuracy to SfM inlier ratio:
\begin{equation}
\label{eq:acc-inlier}
\mathrm{Inlier}_s = y_0 + A\left(1 - \exp\left(-k\,\mathrm{Acc}_s\right)\right),
\end{equation}
equivalently $y = y_0 + A\left(1 - \exp(-k x)\right)$ with $x=\mathrm{Acc}_s$ and $y=\mathrm{Inlier}_s$. We report fitted coefficients and evaluation protocol in Sec.~\ref{sec:experiments}.

%% file: figures/tex/architecture.tex
\begin{figure}[!t]
      \centering
      \includegraphics[width=1.0\linewidth]{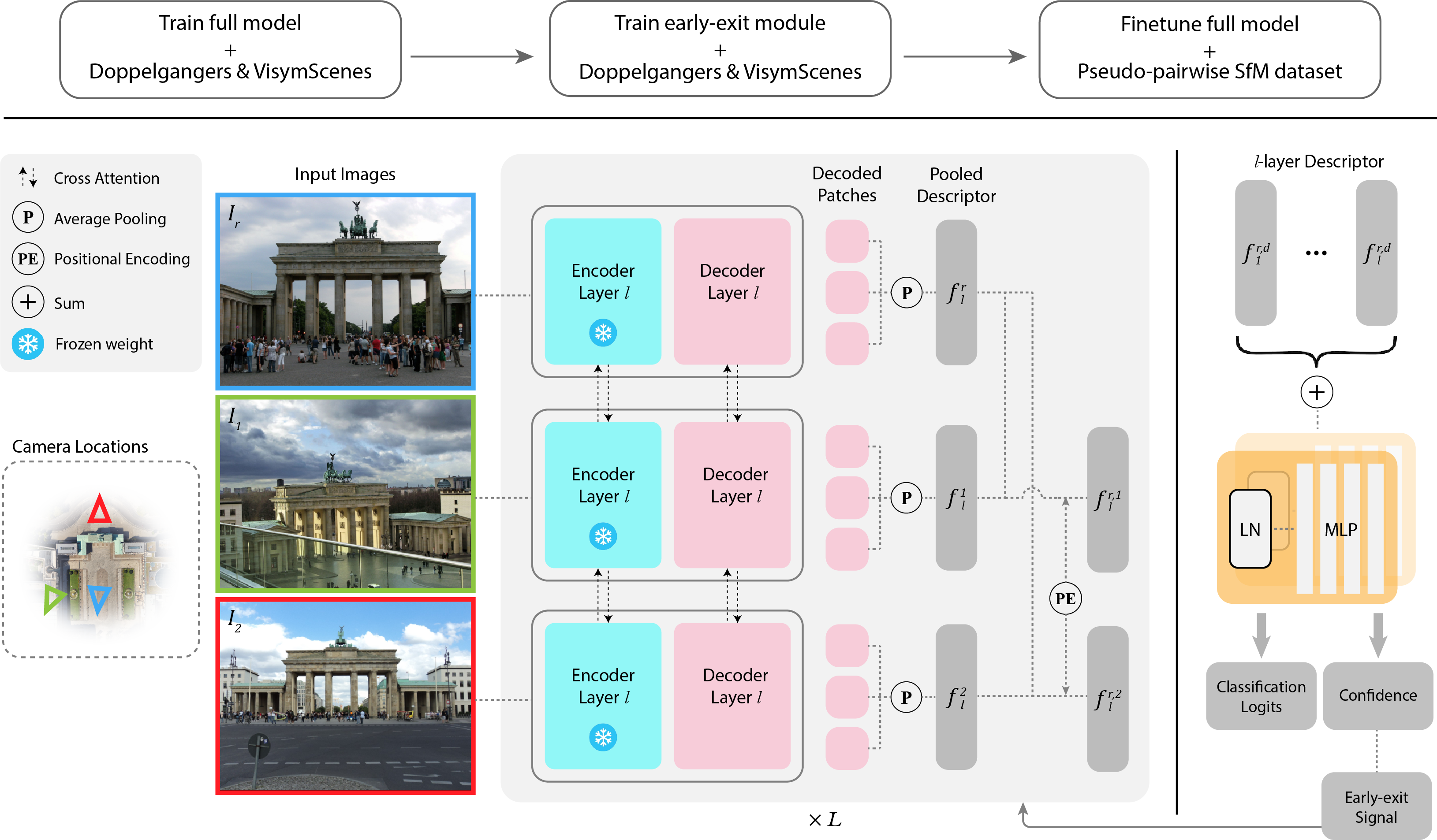}
      \caption{\textbf{Architecture of the proposed pipeline.} Given a reference image ($I_r$) and database images ($I_1, I_2$), our method encodes all inputs and extracts geometry-aware tokens to classify whether each database image corresponds to the same scene and viewpoint as the reference. We freeze the pre-trained VGGT encoder and train a DPT-style decoder to obtain per-patch features, which are then average pooled into global descriptors. Positionally encoded descriptors are fed to an MLP-based \emph{doppelgangers} classifier. To accelerate inference, a confidence-based \emph{early-exit} module halts unnecessary deeper-layer computation. \emph{LN} denotes LayerNorm.} 
      \label{fig:architecture}
\end{figure}

%% file: sec/5_experiments.tex
\section{Experiments}
\label{sec:experiments}

\paragraph{Overview.}
We first describe training data and implementation details, then report quantitative and qualitative results, followed by ablations, a retrieval variant, and downstream SfM evaluations. Plots and fitted curves are provided in this section; formulas are defined in Sec.~\ref{sec:method}.

\subsection{Training Dataset} \label{sec:training-datasets}
We follow~\cite{xiangli2025doppelgangersimprovedvisualdisambiguation} and train our models on a union of \emph{Doppelgangers (DG)}~\cite{cai2023doppelgangerslearningdisambiguateimages}, \emph{MegaDepth}~\cite{MegaDepthLi18}, and \emph{VisymScenes}~\cite{xiangli2025doppelgangersimprovedvisualdisambiguation}. The training split comprises 1{,}184 scenes with 171{,}498 positive and 171{,}508 negative image pairs. To standardize resolution while preserving aspect ratio, images are resized to a fixed width $W\!=\!518$ and height
\(
H \!=\! \left\lfloor H_0 \cdot \frac{518}{W_0} \right\rfloor
\),
where $(H_0, W_0)$ are the original dimensions; all images in a batch share the same $(H, W)$ with padding.
For multi-view training, we transfer pairwise supervision into coarse view groups as described in Sec.~\ref{sec:method-data}. 


\subsection{Implementation Details}
We train with a batch size of $8$ and keep the VGGT encoder \emph{frozen}. For each encoder layer, a lightweight two-layer DPT head produces per-image descriptors, which are fed to the scoring head described in Sec.~\ref{sec:method-module-1}, along with an \emph{early-exit} confidence head. The decoder/channel dimension is set to $1024$ with $8$ attention heads. We use gradient accumulation of $8$ steps for stable training. The learning rate to train the first stage starts at $1\!\times\!10^{-4}$ and follows a stepped schedule down to $1\!\times\!10^{-5}$ by the end of each epoch; for the second stage, the learning rate decays from $1\!\times\!10^{-5}$ to $1\!\times\!10^{-6}$ within a 5000-step schedule. Unless otherwise noted, all experiments are conducted on $2\times$ NVIDIA H100 GPUs. We train the similarity scoring head (Sec.~\ref{sec:method-module-1}) with focal loss (gamma$=1$) and the confidence head (Sec.~\ref{sec:method-module-2}) with binary cross-entropy (BCE) loss.


\subsection{Quantitative Results} \label{sec:quantitative-results}
\input{tables/comparison_ours_vs_doppelgangers}
We quantitatively evaluate our method on the \emph{VisymScenes} test dataset, which provides multiple positive and negative pairs per reference view and has been shown to benefit training for 3D disambiguation~\cite{xiangli2025doppelgangersimprovedvisualdisambiguation}. The test dataset contains $3{,}180$ labeled image pairs drawn from diverse scenes (Tab.~\ref{tab:comparison_ours_vs_doppelgangers}). We report standard metrics: $\text{accuracy} = (T_p + T_n) / \text{total}$, $\text{precision} = T_p / (T_p + F_p)$, $\text{recall} = T_p / (T_p + F_n)$ and $\mathcal{F}_1 = 2 \cdot \text{precision} \cdot \text{recall} / (\text{precision} + \text{recall})$, 
where $T_p, T_n, F_p, F_n$ denote true positives, true negatives, false positives, and false negatives, respectively. Our model outperforms Doppelgangers++ on \emph{VisymScenes} while running faster, but underperforms on the DG++ pairwise benchmark (Tab.~\ref{tab:comparison_ours_vs_doppelgangers}). We suspect this gap is related to supervision noise in \emph{Doppelgangers} labels and additional ambiguity introduced when pairwise labels are propagated to multi-view groups for training and validation; these failure modes are illustrated in Sec.~\ref{sec:limitations}. Ablation studies (Sec.~\ref{sec:ablation}) further analyze the contributions of layer selection and confidence gating. For clarity, the \texttt{Ours} score in Tab.~\ref{tab:comparison_ours_vs_doppelgangers} (0.9441 on \emph{VisymScenes}) corresponds to the full model without early exit, while the higher 0.9503 result is obtained after enabling confidence-driven early exit (Tab.~\ref{tab:confidence_accuracy_vs_time}, $\tau=0.8$).

We also benchmark against the state-of-the-art VPR method SALAD~\cite{Izquierdo_CVPR_2024_SALAD} on the same test set. Because SALAD outputs a static global descriptor for each image, we perform a 200-step threshold sweep over cosine similarity to select the operating point that maximizes $\mathcal{F}_1$ on the test dataset; at the optimal threshold ($0.3171$), SALAD still yields more false positives and false negatives than both Doppelgangers++ and our method (Tab.~\ref{tab:comparison_ours_vs_doppelgangers}). This gap reflects a mismatch in objective: VPR pipelines optimize global descriptor retrieval in 2D and are not explicitly trained to resolve ambiguous 3D geometry. Without modeling multi-view geometric consistency, 2D descriptors can be deceived by visually similar, non-covisible structures prevalent in our benchmark.

\subsection{Qualitative Results} \label{sec:qualitative-results}
\input{figures/tex/comparison-qualitative}
We further compare against Doppelgangers++ and SALAD in Fig.~\ref{fig:qualitative-comparison-results} on challenging scenes. Our approach adapts to illumination and viewpoint changes by efficiently extracting geometry-aware cues from multiple views. It also proves robust to transient distractors (e.g., pedestrians, vehicles, seasonal decorations), focusing instead on stable structural elements such as building facades and storefronts.

Additional qualitative examples and analysis are provided in the supplementary material.



\subsection{Ablation Studies} \label{sec:ablation}
\input{tables/ablation}

\paragraph{Pairwise vs. Multi-view.}
We compare full-model training under pairwise-only supervision versus our multi-view supervision transfer (Sec.~\ref{sec:method-data}) to measure whether multi-view context improves downstream behavior beyond runtime gains. Relative to pairwise training, multi-view training yields consistent gains on broader-range metrics and SfM relevance: AUC@5 improves from $0.690$ to $0.701$ ($+0.011$), AUC@10 from $0.753$ to $0.775$ ($+0.022$), and inlier ratio from $0.7579$ to $0.7698$ ($+0.0119$), while AUC@3 changes only marginally from $0.626$ to $0.624$ ($-0.002$). Overall, the trend indicates that exposing the model to multi-view geometric context improves medium/long-range discrimination and translates to more reliable SfM-oriented filtering.

\subsection{Pseudo-pairwise SfM Dataset} \label{sec:pp-vs-sfm}

\input{figures/tex/pseudopairwise}
\input{figures/tex/sfm-qualitative}

We use the pseudo-pairwise SfM dataset processed from \emph{AerialMegaDepth} in two ways. During validation, we use pseudo-pairwise accuracy as a lightweight proxy for downstream SfM quality, which avoids running expensive full SfM evaluation at every checkpoint. After convergence on \emph{Doppelgangers} and \emph{VisymScenes}, we fine-tune on this pseudo-pairwise dataset to improve training stability and downstream SfM behavior on the test set. 

\paragraph{Correlation between pseudo-pairwise accuracy \& SfM metrics.}
We analyze how pseudo-pairwise accuracy predicts downstream SfM accuracy on processed scenes (Sec.~\ref{sec:method-sfm}). We evaluate 44 model checkpoints and fit a saturating exponential curve to capture the plateau behavior when pseudo accuracy saturates while SfM metrics fluctuate within a high-value regime. The fitted parameters are $y_0=0.572$, $A=0.412$, and $k=3.51$ with $\mathrm{MSE}=9.56\times 10^{-3}$ for inlier ratio, and $y_0=0.586$, $A=0.412$, and $k=3.11$ with $\mathrm{MSE}=1.23\times 10^{-2}$ for AUC@3. As shown in Fig.~\ref{fig:pseudopairwise-correlation}, variance shrinks as SfM inlier ratio increases. Therefore, when pseudo accuracy is stably high ($\geq 0.9$), SfM performance also tends to become stable after pseudo-pairwise fine-tuning.

\paragraph{Improvement in training stability.} We quantify stability over checkpoints from iterations 38k to 112k (saved every 2k steps), comparing models with and without pseudo-pairwise finetuning on \emph{AerialMegaDepth}. To avoid data leakage, we hold out scenes for pseudo-pairwise fine-tuning and report SfM metrics only on disjoint test scenes. Finetuning increases the mean number of inlier cameras from $241.67$ to $266.06$ and reduces its deviation from $45.11$ to $25.46$, indicating substantially more stable reconstructions. Translation and rotation values denote camera-pose errors, i.e., differences between reconstructed camera poses and ground-truth poses; these also improve in mean (translation: $2581.44\!\rightarrow\!448.77$; rotation: $5.96\!\rightarrow\!3.70$), with much lower translation deviation ($3864.74\!\rightarrow\!878.97$) and a slight increase in rotation deviation ($3.26\!\rightarrow\!3.53$). For camera-pose AUC, trends are mixed: AUC@3 improves slightly ($0.3167\!\rightarrow\!0.3238$), while AUC@5 and AUC@10 decrease ($0.5010\!\rightarrow\!0.4699$ and $0.7062\!\rightarrow\!0.6082$), albeit with lower deviations after finetuning. Overall, pseudo-pairwise finetuning provides markedly better training stability, especially for inlier count and translation consistency.

\subsection{Structure-from-motion Results} \label{sec:sfm-results}

\input{tables/rebuttal-sfm-quantitative}
We evaluate our method as a view-graph pruning module before SfM using GLOMAP on \emph{AerialMegaDepth}~\cite{MegaDepthLi18,vuong2025aerialmegadepth}. Following Sec.~\ref{sec:pp-vs-sfm}, we use only MegaDepth ground-level images as model input and use aerial views only as reference-pose supervision for metric computation. For each reconstruction, we align predicted and reference cameras with RANSAC. Higher inlier ratio (reconstructed cameras / input cameras after pruning) and higher pose AUC indicate better geometric accuracy. We use the same decision threshold as DG++ ($\tau{=}0.8$).

%% file: tables/comparison_ours_vs_doppelgangers.tex
\begin{table}[!ht]
    \centering
    \caption{\textbf{Comparison of image disambiguation methods on pairwise test datasets.}}
    \resizebox{1.0\columnwidth}{!}{

    \begin{tabular}{l|cc|cc|cc|cc|cc}
    \toprule
                                & \multicolumn{2}{c|}{Accuracy$\uparrow$} & \multicolumn{2}{c|}{Precision$\uparrow$} & \multicolumn{2}{c|}{Recall$\uparrow$} & \multicolumn{2}{c|}{$\mathcal{F}_1\uparrow$} & \multicolumn{2}{c}{Inference time (msec)$\downarrow$}\\
                                & VisymScenes & DG++ & VisymScenes & DG++ & VisymScenes & DG++ & VisymScenes & DG++ & \\
    \midrule
    SALAD                       & 0.8104 & 0.6233 & 0.7695 & 0.5894 & 0.8862 & 0.8128 & 0.8237 & 0.6833 & \multicolumn{2}{c}{9.87}\\
    Doppelgangers++             & 0.9355 & \textbf{0.937} & 0.9095 & \textbf{0.970} & 0.9673 & 0.901 & 0.9375 & \textbf{0.934} & \multicolumn{2}{c}{165}\\
    Ours                        & \textbf{0.9441} & 0.889 & \textbf{0.9215} & 0.877 & \textbf{0.9712} & \textbf{0.903} & \textbf{0.9457} & 0.890 & \multicolumn{2}{c}{103.92} \\
    \bottomrule
    \end{tabular}
    }
    \label{tab:comparison_ours_vs_doppelgangers}
\end{table}

%% file: figures/tex/comparison-qualitative.tex
\begin{figure}[htb]
      \centering
      \includegraphics[width=1.0\linewidth]{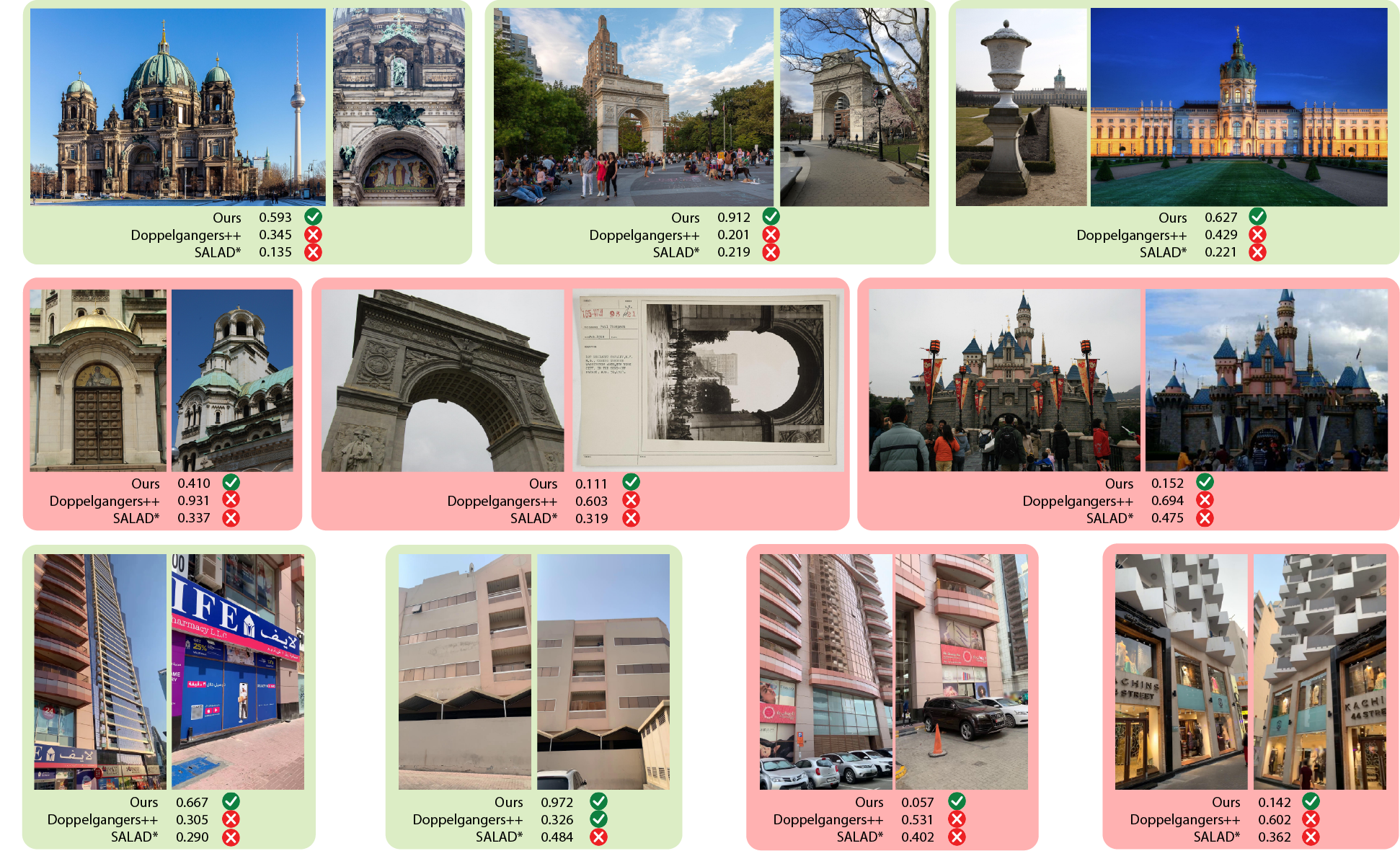}
      \caption{\textbf{Qualitative comparison on \emph{Doppelgangers} and \emph{VisymScenes}~\cite{cai2023doppelgangerslearningdisambiguateimages,xiangli2025doppelgangersimprovedvisualdisambiguation}.} The predicted classification probabilities of \emph{ours} and \emph{Doppelgangers++} are thresholded at $0.5$. Because SALAD~\cite{Izquierdo_CVPR_2024_SALAD} outputs static global descriptors, \emph{SALAD*} denotes cosine similarity thresholded at $0.3171$, which gives the best $\mathcal{F}_1$ score (Sec.~\ref{sec:quantitative-results}). The legend shown in the first top-left negative example applies to all examples; each tile follows the same annotation scheme. \textcolor{red}{Red}-boxed examples are false pairs (different scene/view), and \textcolor{green}{green}-boxed examples are true pairs (same scene/view).}
      \label{fig:qualitative-comparison-results}
      \vspace{-0.2cm}
\end{figure}

%% file: tables/ablation.tex
\begin{table}
    \centering
    \begin{minipage}[t]{0.49\linewidth}
    \centering
    \caption{\textbf{Ablation study on pooling and fine-tuning variants on the \emph{VisymScenes} test set~\cite{xiangli2025doppelgangersimprovedvisualdisambiguation}.}}
    \resizebox{\linewidth}{!}{
    \begin{tabular}{l|llll}
    \toprule
                                & Accuracy$\uparrow$    & Precision$\uparrow$       & Recall$\uparrow$  & $\mathcal{F}_1\uparrow$          \\ 
    \midrule
    Ours w/ AttnPool            & 0.9378                & 0.9156                    & 0.9649            & 0.9396                \\ 
    Ours w/ GeMPool             & 0.9039                & 0.8790                    & 0.9373            & 0.9072                \\ 
    Ours (retrieval)            & 0.9089                & 0.8766                    & 0.9524            & 0.8969                \\ 
    Ours w/ finetune            & \textbf{0.9522}       & \textbf{0.9413}           & 0.9649            & \textbf{0.9530}       \\ 
    Ours                        & \underline{0.9441}    & \underline{0.9215}        & \textbf{0.9712}   & \underline{0.9457}    \\ 
    \bottomrule
    \end{tabular}
    }
    \label{tab:ablation-studies}
    \end{minipage}\hfill
    \begin{minipage}[t]{0.49\linewidth}
    \centering
    \caption{\textbf{Ablation study on confidence threshold: accuracy vs. inference time on the \emph{VisymScenes} test set.}}
    \resizebox{\linewidth}{!}{%
    \begin{tabular}{l|llll}
    \toprule
    Confidence Threshold & 0.5    & 0.8    & 0.9    & 0.95  \\ 
    \midrule
    Inference Time (msec)  & 99.56  & 103.92  & 104.42  & 103.69   \\ 
    Accuracy             & 0.8676 & 0.9503 & 0.9500 & 0.9447  \\ 
    \bottomrule
    \end{tabular}
    }
    \label{tab:confidence_accuracy_vs_time}
    \end{minipage}
\end{table}

%% file: figures/tex/pseudopairwise.tex
\begin{figure}[!t]
      \centering
      \begin{subfigure}[t]{0.48\linewidth}
            \centering
            \includegraphics[width=\linewidth]{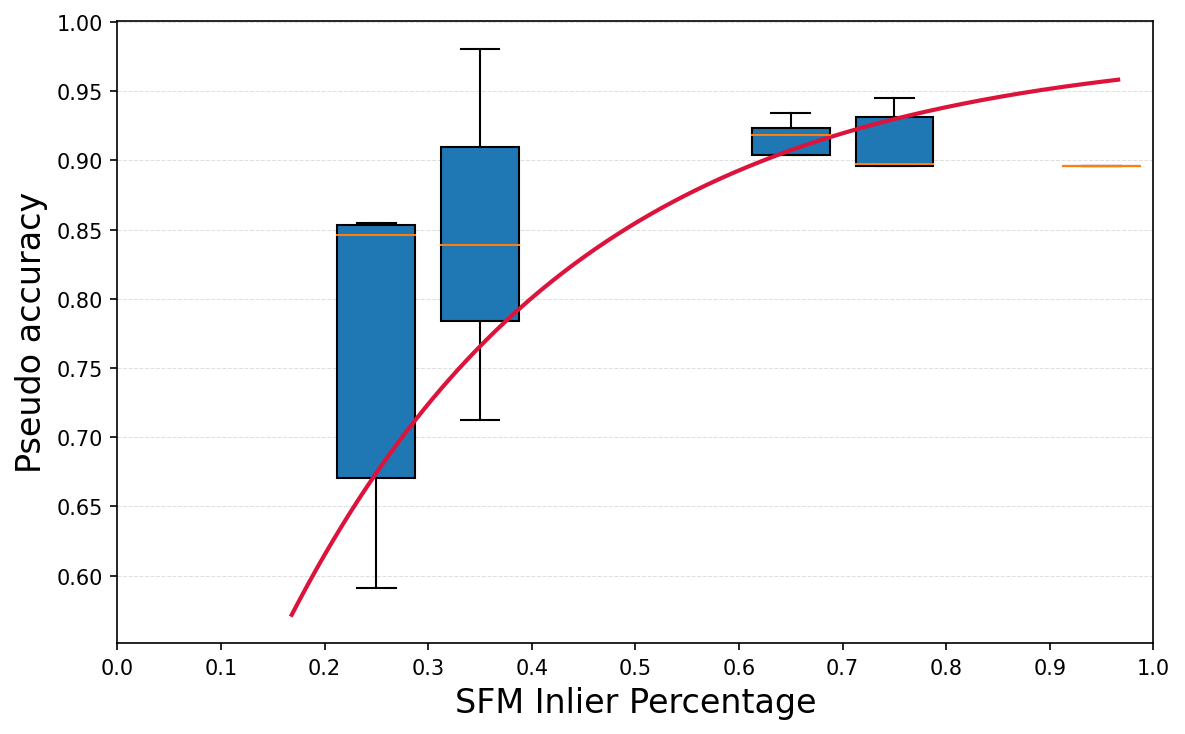}
            \caption{SfM inlier ratio.}
      \end{subfigure}
      \hfill
      \begin{subfigure}[t]{0.48\linewidth}
            \centering
            \includegraphics[width=\linewidth]{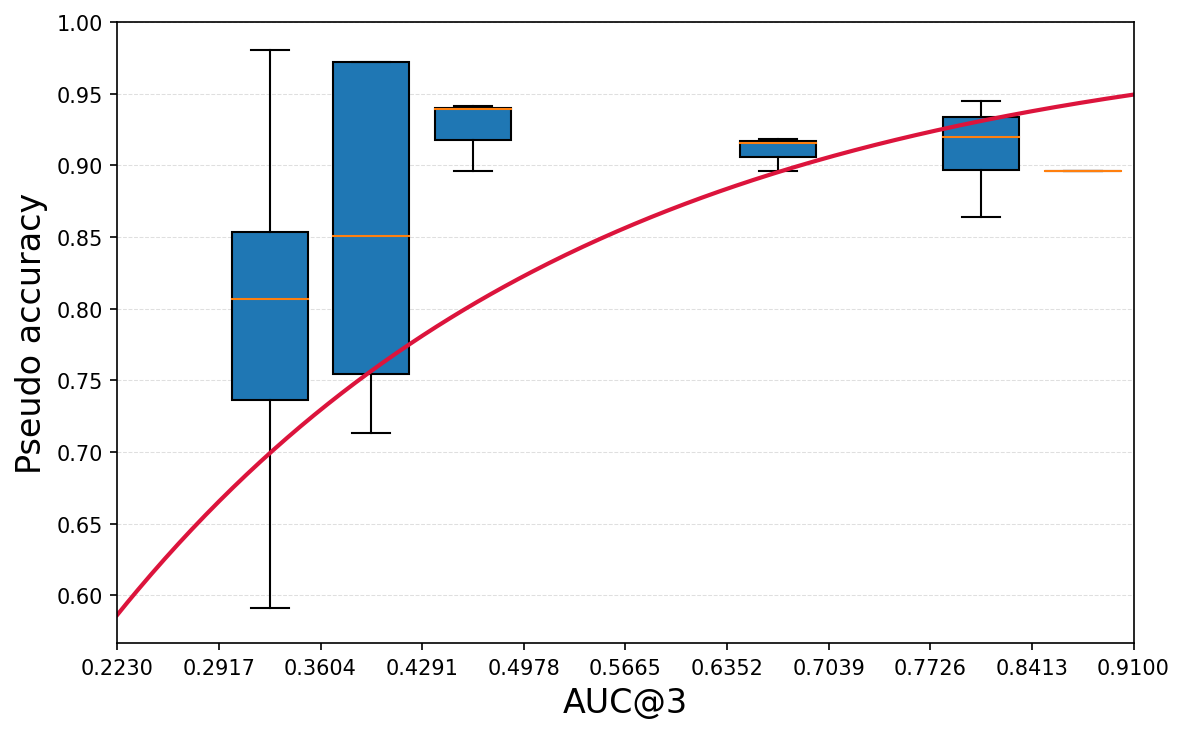}
            \caption{AUC@3.}
      \end{subfigure}
      \caption{\textbf{Correlation between pseudo-pairwise accuracy and downstream SfM metrics.}}
      \label{fig:pseudopairwise-correlation}
\end{figure}

%% file: figures/tex/sfm-qualitative.tex
\begin{figure}[!t]
      \centering
      \includegraphics[width=.95\linewidth]{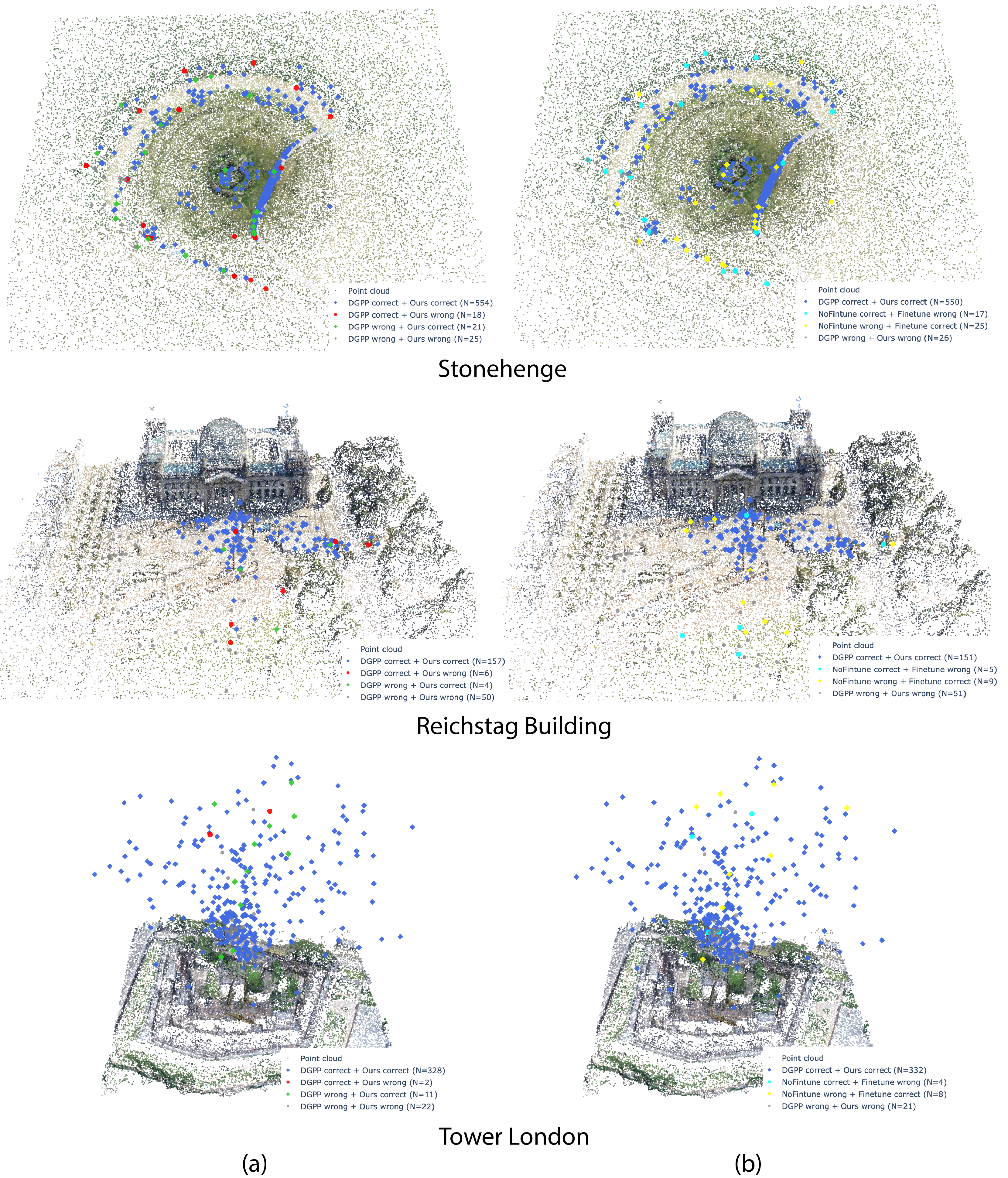}
      \caption{\textbf{Qualitative results on SfM using \emph{AerialMegaDepth}.} For visualization, we clip the point cloud to $[-150, 150]^2$ on the ground plane to focus on the scene center; this removes some points and cameras.}
      \label{fig:sfm-qualitative-results}
\end{figure}

%% file: tables/rebuttal-sfm-quantitative.tex
\begin{table}[h]
    \centering
    \caption{\textbf{Quantitative SfM evaluation on \emph{AerialMegaDepth}.}}
    \resizebox{1.0\columnwidth}{!}{%
\begin{tabular}{l|ll|ll|ll|ll}
\toprule
                       & \multicolumn{2}{l|}{AUC@3}     & \multicolumn{2}{l|}{AUC@5}    & \multicolumn{2}{l|}{AUC@10}   & \multicolumn{2}{l}{Inlier ratio}     \\ \cline{2-9} 
                       & DG++           & Ours          & DG++          & Ours          & DG++          & Ours          & DG++             & Ours               \\ \midrule
Big Ben                & 0.326          & \textbf{0.339}& 0.455         & \textbf{0.537}& 0.511         & \textbf{0.747}& 252/748          & \textbf{256/730}   \\
Louvre Pyramid         & 0.830          & \textbf{0.852}& 0.886         & \textbf{0.896}& 0.928         & \textbf{0.930}& 207/296          & \textbf{227/297}   \\
Phatheon Paris         & \textbf{0.912} & 0.910         & 0.929         & 0.929         & \textbf{0.944}& 0.942         & 491/537          & \textbf{493/536}   \\
Ponte Di Rialto        & 0.068          & \textbf{0.442}& 0.246         & \textbf{0.466}& 0.438         & \textbf{0.483}& 161/329          & \textbf{173/369}   \\
Reichstag Building     & 0.720          & \textbf{0.723}& 0.832         & 0.832         & \textbf{0.916}& 0.914         & \textbf{163/222} & 161/221            \\
Sacre Cœur Basilica    & \textbf{0.811} & 0.797         & \textbf{0.884}& 0.875         & \textbf{0.940}& 0.935         & 561/699          & \textbf{570/696}   \\
St. Stephens Cathedral & 0.852          & \textbf{0.855}& 0.910         & \textbf{0.911}& 0.953         & 0.953         & \textbf{647/699} & 642/698            \\
St. Vitus Cathedral    & 0.693          & \textbf{0.707}& 0.815         & \textbf{0.823}& 0.907         & \textbf{0.910}& 482/500          & \textbf{483/500}   \\
Tower London           & 0.493          & \textbf{0.496}& 0.511         & \textbf{0.513}& 0.527         & 0.527         & 469/561          & \textbf{482/562}   \\
\bottomrule
\end{tabular}
}
\label{tab:rebuttal-sfm-quantitative}
\end{table}

%% file: sec/6_discussion.tex
\section{Conclusion}
\label{sec:conclusion}

In this work, we addressed multi-view 3D disambiguation with a confidence-driven early-exit framework built on a strong feed-forward backbone. By leveraging selective intermediate-layer tokens from a frozen VGGT encoder and coupling them with a lightweight decoder-classifier, we show that many scenes can be disambiguated confidently before the deepest layers. This reduces computation and often improves accuracy by avoiding overly invariant deep representations that can reintroduce ambiguity. Our approach complements both 2D retrieval pipelines and pairwise 3D-feature methods, offering competitive disambiguation accuracy with stronger scalability and runtime efficiency in multi-view settings. Future work includes tighter integration with downstream SfM/MVS pipelines for closed-loop filtering, as well as extensions to streaming video and active view selection. We also plan to audit and correct annotation errors in \emph{Doppelgangers}, and to release improved multi-view disambiguation datasets for the community.

%% file: sec/X_suppl.tex
\clearpage
\setcounter{page}{1}

\paragraph{Overview.} 
In this supplementary material, we present our solution to a key dataset bottleneck (Sec.~\ref{sec:design-choice}), additional qualitative results, more SfM evaluation, and extra ablations (Sec.~\ref{sec:supp-qual-ablation}), additional implementation details (Sec.~\ref{sec:impl-details}), and a discussion of failure cases (Sec.~\ref{sec:limitations}).

\section{Permutation-invariant Classification} \label{sec:design-choice}

We observe that classification accuracy degrades significantly when the input image order is shuffled at test time. A likely reason is that some images are more often used as reference views during dataset construction. To mitigate this issue, we enforce permutation invariance in three ways. First, during sampling we randomly swap the two images in each pair with probability $0.5$, exposing the model to both orderings throughout training. Second, we construct a symmetric pair feature as in Eq.~\ref{eq:pair-feat}; specifically, absolute difference and element-wise product are permutation invariant. Third, instead of attention pooling (which can implicitly encode ordering), we use permutation-invariant average pooling over patches to obtain frame descriptors.

\section{Additional Qualitative and Ablation Results} \label{sec:supp-qual-ablation}

\subsection{Additional Qualitative Results}
\input{figures/tex/qualitative}

We evaluate qualitatively on the \emph{Doppelgangers} test set~\cite{cai2023doppelgangerslearningdisambiguateimages}. Fig.~\ref{fig:qualitative-results} visualizes classification for a reference image with two retrieved positives and two retrieved negatives. For each pair, we report predicted confidence, the early-exit layer where inference halts, and the corresponding score (thresholded at $\theta$; Sec.~\ref{sec:method-module-1}). We observe that deeper layers generally yield higher confidence and stronger scores.

\subsection{More SfM Evaluation}
\input{figures/tex/supplementary-sfm}
In Fig.~\ref{fig:supp-sfm}, we show more structure-from-motion results on \emph{AerialMegaDepth} scenes that complement Fig.~\ref{fig:sfm-qualitative-results} in the main paper. Again, we clip the point cloud to $[-150, 150]^2$ on the ground plane to focus on the scene center, and this may remove some points and cameras for demonstration purposes.

In the supplementary ZIP file, we provide the HTML used to generate Fig.~\ref{fig:sfm-qualitative-results} and Fig.~\ref{fig:supp-sfm}. Due to upload space limits, we include only a subset of scenes. You may hover the cursor around the point cloud and click on cameras to open and inspect the corresponding images.

\subsection{Additional Ablation Results}
\input{figures/tex/encoder-layer-attention}
\input{tables/fixed-layers-accuracy}
\input{tables/rebuttal-dpt-vs-encoder}

\paragraph{Choosing encoder layers via attention.}
We compare fixed-layer selection against (i) using all layers and (ii) our confidence-driven early exit. The VGGT paper aggregates fixed layers $[4,7,11,23]$ for downstream decoding; however, in our classification setting this choice is prone to overfitting, with training accuracy rising while test accuracy lagging. Tab.~\ref{tab:fixed-layers-accuracy} shows that using only early layers (e.g., $[4]$) yields poor generalization, while $[4,11,17]$ improves convergence yet remains inferior to using all layers or our adaptive policy. Fig.~\ref{fig:encoder-layer-attention} visualizes attention maps across depth: we observe extended stretches of qualitatively similar patterns, with notably distinct behaviors emerging around layers $4$, $17$, and $23$. This supports our design to (a) expose the classifier to multiple depths during training and (b) allow the model to \emph{adapt} the effective depth at inference.

\paragraph{DPT-style decoder vs. encoder-only variant.}
We compare two heads on \emph{AerialMegaDepth}: a DPT-style decoder and a lighter encoder-only global descriptor head (Tab.~\ref{tab:rebuttal-dpt-vs-encoder}). Their headline SfM metrics are very close, with the encoder-only variant being slightly faster. We retain the DPT-style design because it preserves patch-/image-space outputs that are useful for follow-up geometry tasks (e.g., dense correspondence reasoning), while maintaining competitive downstream SfM performance.

\paragraph{Confidence thresholds vs.\ accuracy and time.}
We study how the confidence threshold $\tau \in \{0.50, 0.80, 0.90, 0.95\}$ trades off accuracy and latency (Tab.~\ref{tab:confidence_accuracy_vs_time}) on the \emph{VisymScenes} test set. Inference time is reported as average wall-clock latency per image group. Fig.~\ref{fig:confThreshold_vs_used_layer_dist} shows how $\tau$ affects the earliest selected layer: across thresholds, the confidence head selects pre-final layers for a substantial fraction of pairs, indicating significant opportunity for early halting while preserving strong accuracy around $\tau=0.8$.
\input{figures/tex/confThreshold_vs_used_layer_dist}

\paragraph{More SfM metrics on fine-tune vs. skipping}
\input{figures/tex/finetune_stability}

In Fig.~\ref{fig:finetune-stability}, we show detailed SfM metrics that demonstrate the effectiveness of fine-tuning our multi-view model on the proposed pseudo-pairwise dataset. By testing on held-out scenes (i.e., not used in training or fine-tuning), we observe improved stability on the standard SfM task in translation and rotation errors, compared to skipping fine-tuning. We also observe improved test metrics, such as higher inlier ratio and a larger number of inlier cameras. Although there is no conspicuous improvement in AUC@X, the quantitative results in Sec.~\ref{sec:pp-vs-sfm} of the main paper show that this extra step yields comparable overall results.

\section{Implementation Details} \label{sec:impl-details}
\paragraph{VGGT backbone aggregator.}
We reuse the VGGT encoder as a frozen backbone. Given a batch of $B$ images and $S$ frames per scene, the aggregator produces a list of per-layer token tensors 
\[
\{\mathbf{X}^{(l)}\}_{l=1}^{L}, \quad \mathbf{X}^{(l)} \in \mathbb{R}^{B \times S \times P \times C},
\]
and a patch start index indicating the token range corresponding to image patches. We use up to layer $23$ when constructing the head inputs. We observe that training can become unstable when deeper-layer encodings are added indiscriminately. Therefore, we early-stop stage 1 and then train stage 2 (the early-exit module). Using earlier encoding layers for easier cases yields more stable performance.

\paragraph{Classification head.}
For each selected encoder layer index $l$, patch tokens are first projected to a decoder dimension via a per-frame linear layer, augmented with a learned layer positional embedding. We then apply a stack of PyTorch \texttt{TransformerDecoderLayer}s per selected layer, with positional encoding over patches. We decode tokens from each layer independently following VGGT.
We then reduce the layer dimension by summation followed by LayerNorm, and reduce the patch dimension with average pooling. A final MLP classifier maps each frame descriptor to logits over two classes (false and true).

Because the training dataset provides only pairwise images, the classification head outputs $y\in \mathbb{R}^{B \times 2 \times 2}$, corresponding to two frames and two classes. At test time, when more than two images are provided, the same head can optionally output $y\in \mathbb{R}^{B \times (S-1) \times 2 \times 2}$, representing anchor-versus-others pairwise predictions.

\section{Limitations} \label{sec:limitations}
\input{figures/tex/failure-cases}
We present failure cases in Fig.~\ref{fig:failure-cases}. Our pipeline is weaker in four scenarios. First, large domain shifts between input images (e.g., capture time, season, or color processing) reduce robustness, likely because VGGT was trained primarily on static-scene reconstruction within narrower domains. Fine-tuning the VGGT encoder on this data can partially mitigate this issue, as shown in the main paper. Second, if one image in a pair is an extreme zoom-in, finding reliable correspondences becomes difficult. Third, ground-truth labels can be ambiguous. As described in~\cite{cai2023doppelgangerslearningdisambiguateimages}, \emph{Doppelgangers} relies on coarse directional labels (e.g., north/south), without strict geometric rules such as treating viewpoint differences greater than $\ang{90}$ as false pairs. Consequently, in the third row (\emph{Vaguely Labeled}), two pairs with similar visual offsets (about $\ang{45}$) receive different labels. Fourth, some pairs are visually nearly indistinguishable and lack additional localization cues; this becomes especially challenging when horizontal flips are used in augmentation, as a pair labeled true before flipping can become effectively false after flipping.


%% file: figures/tex/qualitative.tex
\begin{figure}[htb]
      \centering
      \includegraphics[width=.95\linewidth]{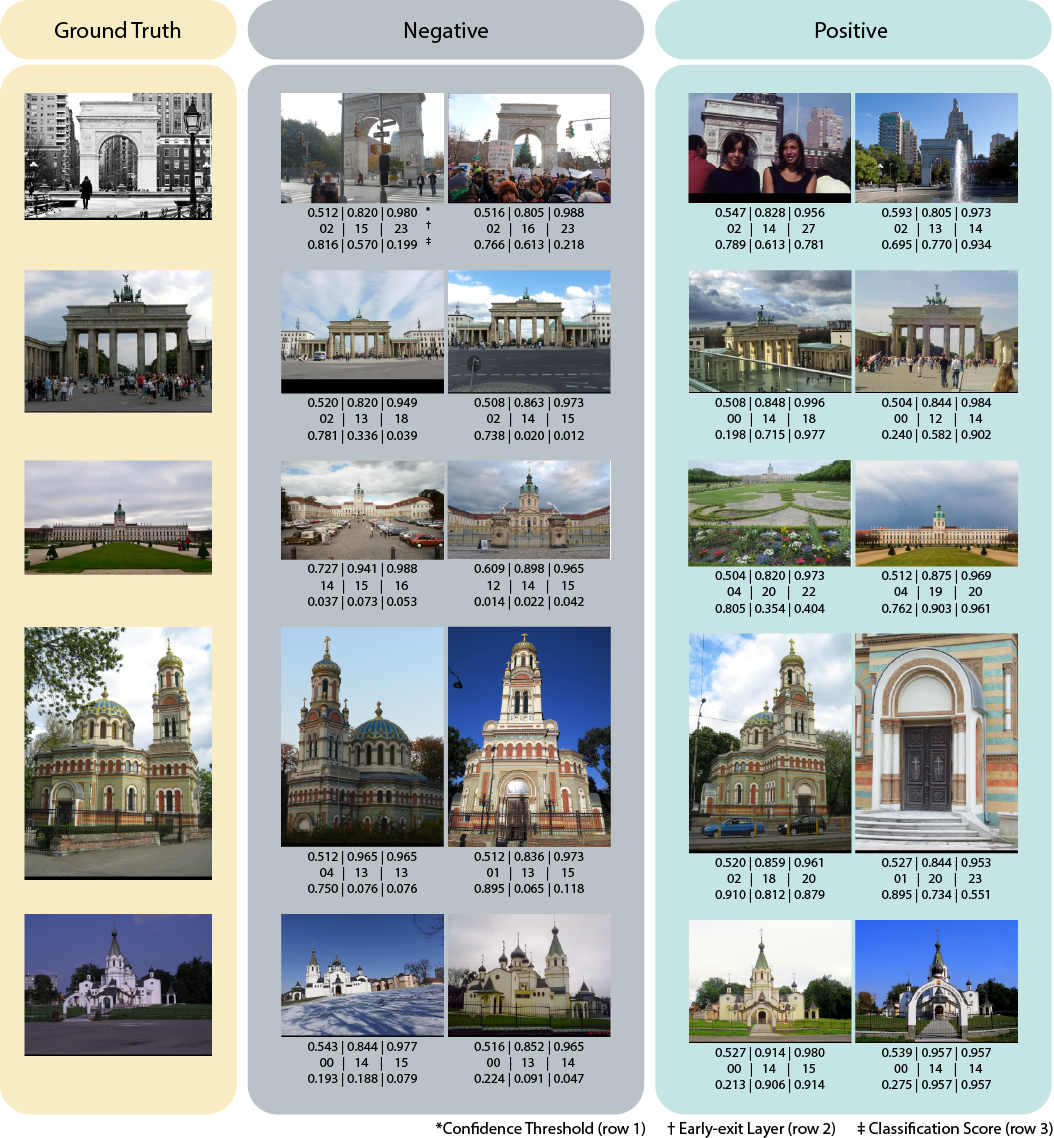}
      \caption{\textbf{Qualitative results on the \emph{Doppelgangers} dataset~\cite{cai2023doppelgangerslearningdisambiguateimages}.} Classification scores are thresholded at $0.5$: scores $<0.5$ are classified as false pairs, and scores $\ge 0.5$ as true pairs.} 
      \label{fig:qualitative-results}
\end{figure}

%% file: figures/tex/supplementary-sfm.tex
\begin{figure}[h]
      \centering
      \includegraphics[height=1.0\textheight]{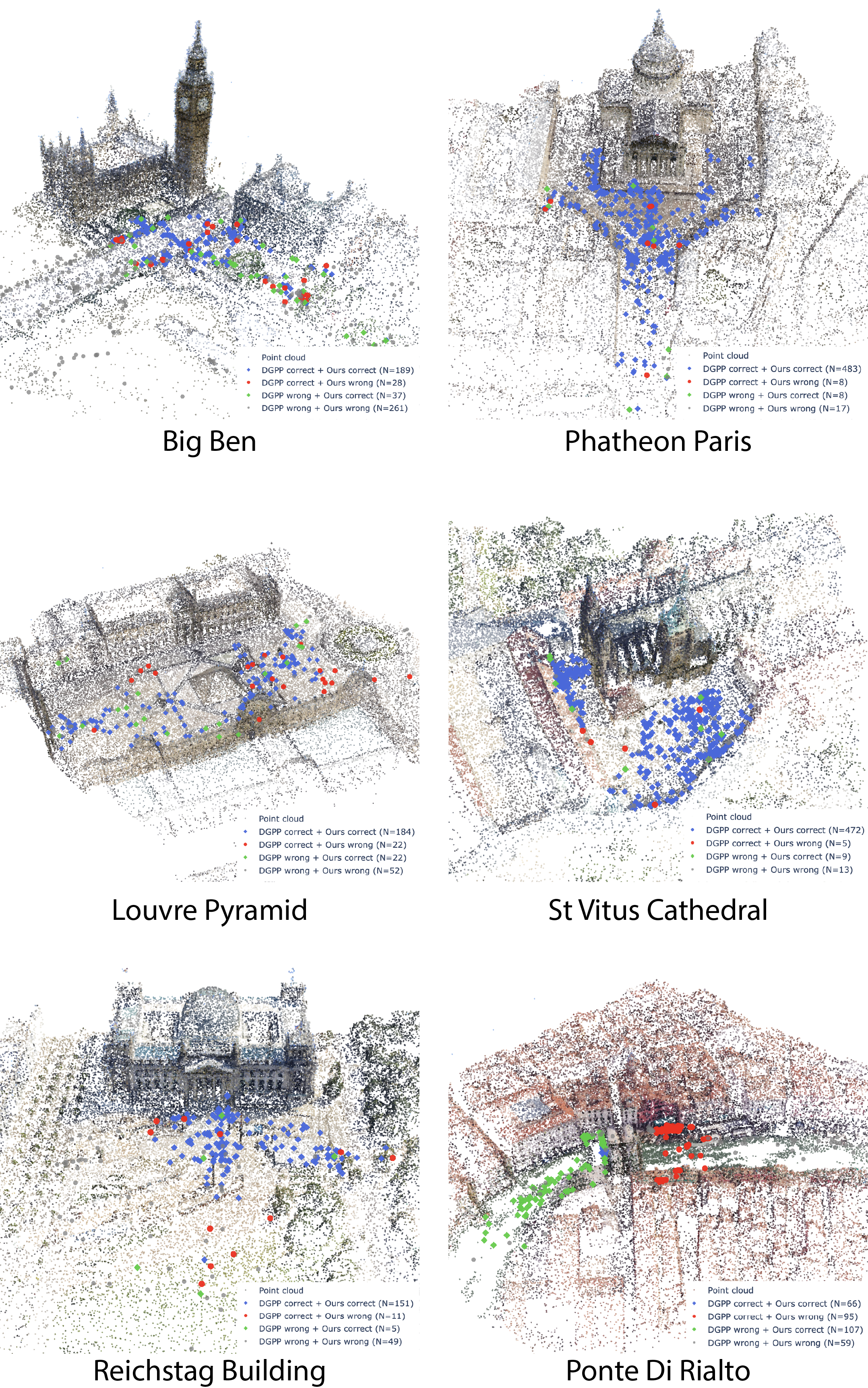}
      \caption{\textbf{More SfM testing on \emph{AerialMegaDepth} scenes.}} 
      \label{fig:supp-sfm}
\end{figure}

%% file: figures/tex/encoder-layer-attention.tex
\begin{figure}[htb]
      \centering
      \includegraphics[width=1.0\linewidth]{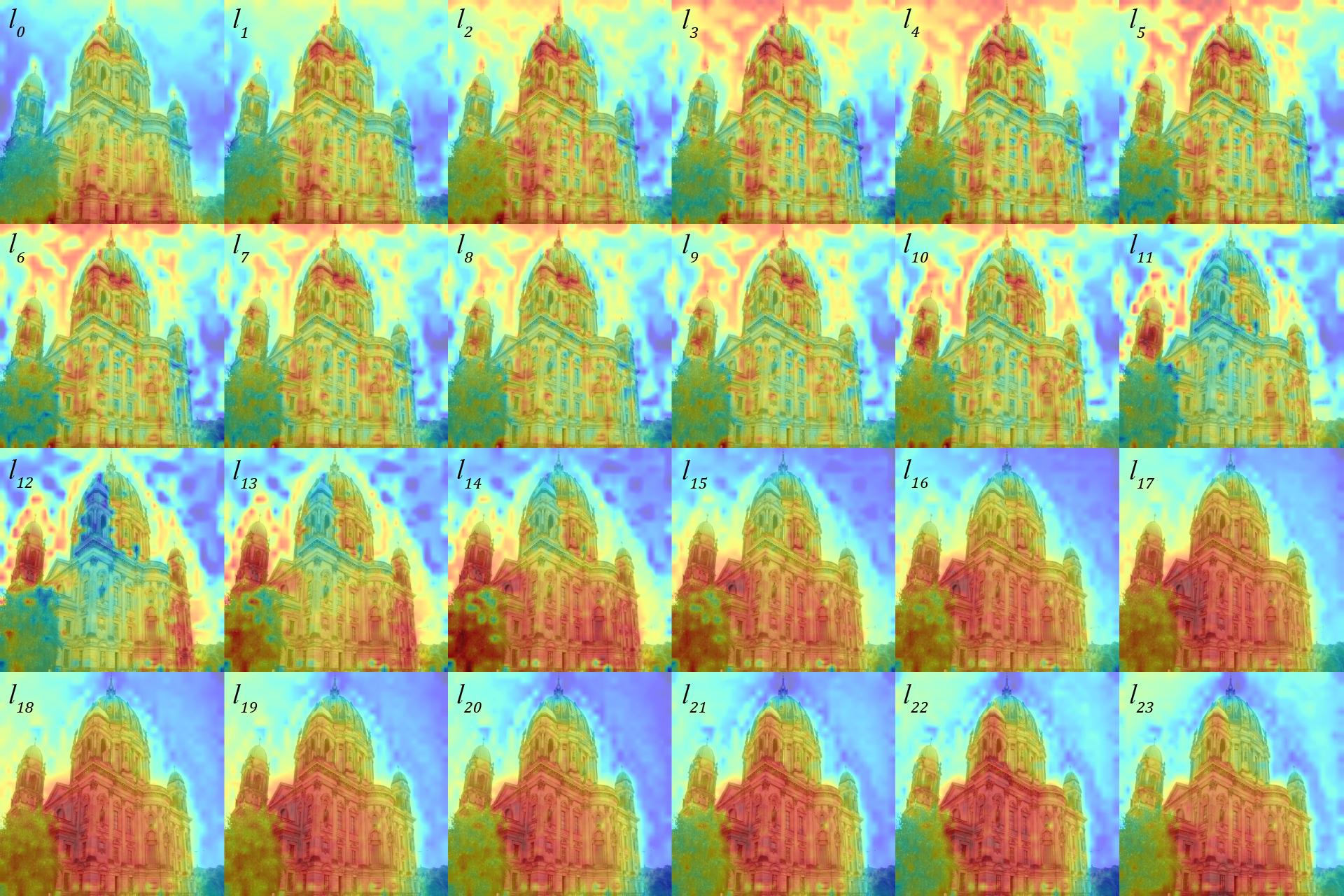}
      \caption{\textbf{Heatmaps of encoder-layer attention.} We average output patch tokens at each encoder layer and upsample them via bilinear interpolation. The heatmaps show similar attention patterns across $l_0,\ldots,l_{11}$.} 
      \label{fig:encoder-layer-attention}
\end{figure}

%% file: tables/fixed-layers-accuracy.tex
\begin{table}[!ht]
    \centering
    \caption{\textbf{Ablation study on encoder layer selection for classification accuracy.}}
    \resizebox{0.5\columnwidth}{!}{%
    \begin{tabular}{l|llll}
    \toprule
    Used Layers       & $[4]$  & $[4, 11]$ & $[4, 11, 17]$  & $[4, 11, 17, 23]$ \\
    \midrule
    Training Accuracy & 0.9547 & 0.8929    & 0.9846         & 0.8904            \\
    Test Accuracy     & 0.5245 & 0.6570    & 0.7028         & 0.7598             \\
    \bottomrule
    \end{tabular}
    }
    \label{tab:fixed-layers-accuracy}
\end{table}

%% file: tables/rebuttal-dpt-vs-encoder.tex
\begin{table}[h]
    \centering
    \caption{\textbf{DPT-style decoder vs. transformer encoder (averaged on \emph{AerialMegaDepth}).}}
    \resizebox{0.6\columnwidth}{!}{%
\begin{tabular}{llllll}
\toprule
        & AUC@3 & AUC@5 & AUC@10 & Inlier ratio & Inference speed      \\
\midrule
DPT     & 0.640 & 0.723 & 0.801  & 0.7369        & 3.38 sec/it          \\
Encoder & 0.643 & 0.726 & 0.802  & 0.7360        & 3.10 sec/it          \\
\bottomrule
\end{tabular}
}
\label{tab:rebuttal-dpt-vs-encoder}
\end{table}

%% file: figures/tex/confThreshold_vs_used_layer_dist.tex
\begin{figure}[htb]
      \centering
      \includegraphics[width=.65\linewidth]{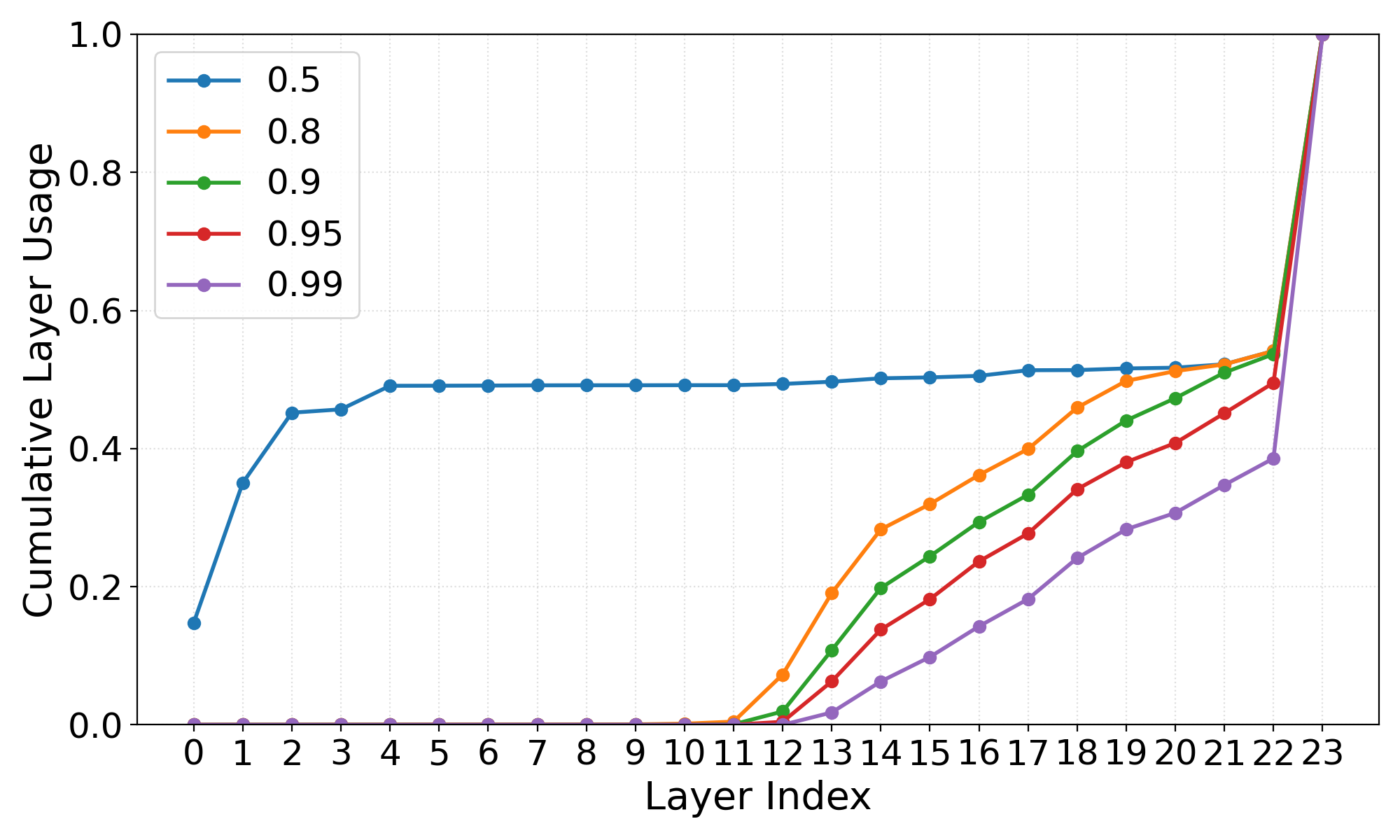}
      \caption{\textbf{Frequency of earliest layers meeting the confidence threshold.} Statistics are computed on the \emph{VisymScenes} test set~\cite{xiangli2025doppelgangersimprovedvisualdisambiguation}.} 
      \label{fig:confThreshold_vs_used_layer_dist}
      \vspace{-0.5cm}
\end{figure}

%% file: figures/tex/finetune_stability.tex
\begin{figure}[htb]
      \centering
      \includegraphics[width=1.0\linewidth]{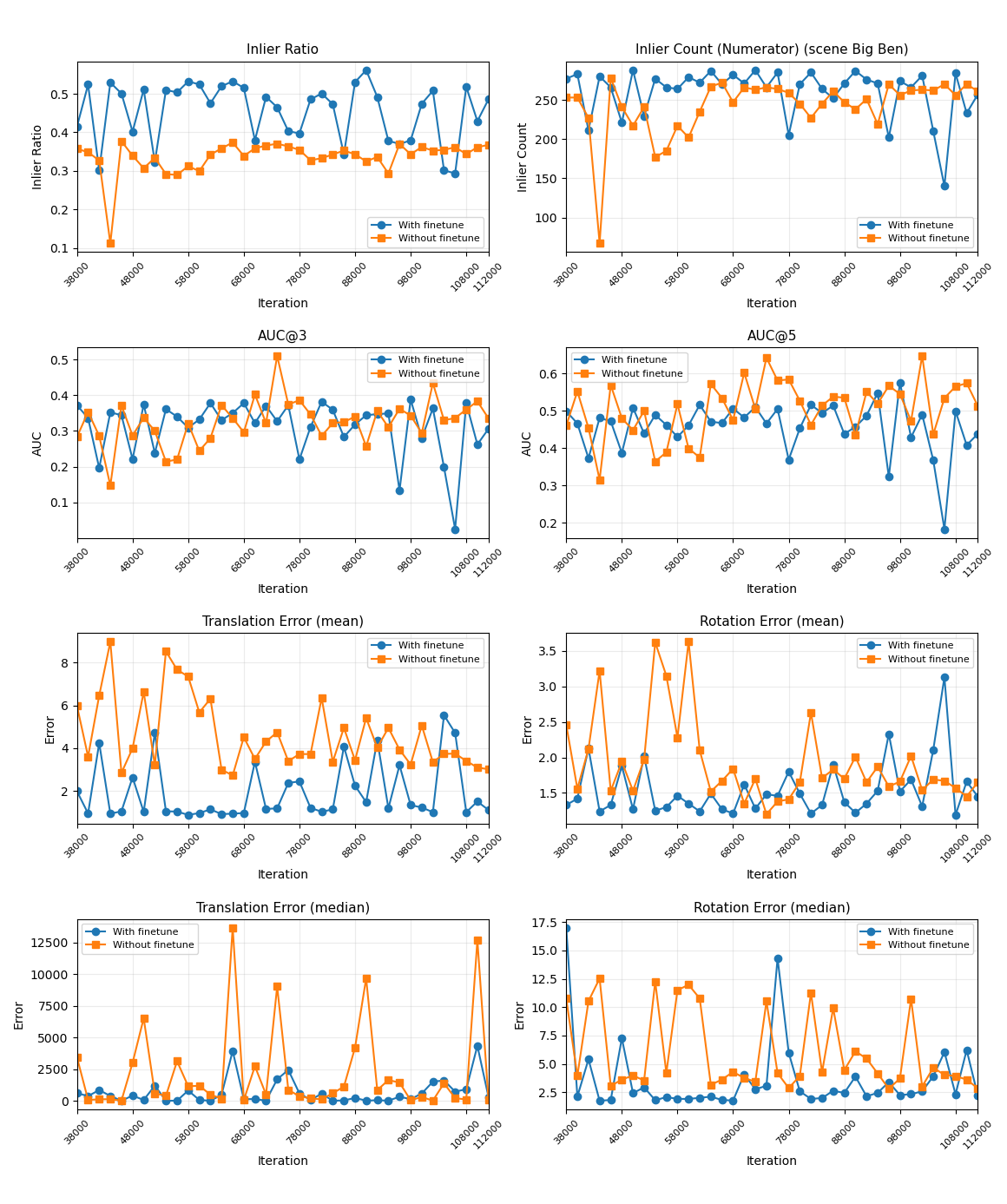}
      \caption{\textbf{Fine-tuning with our pseudo-pairwise dataset vs. skipping, on standard SfM metrics.}} 
      \label{fig:finetune-stability}
\end{figure}

%% file: figures/tex/failure-cases.tex
\begin{figure}[htb]
      \centering
      \includegraphics[width=1.0\linewidth]{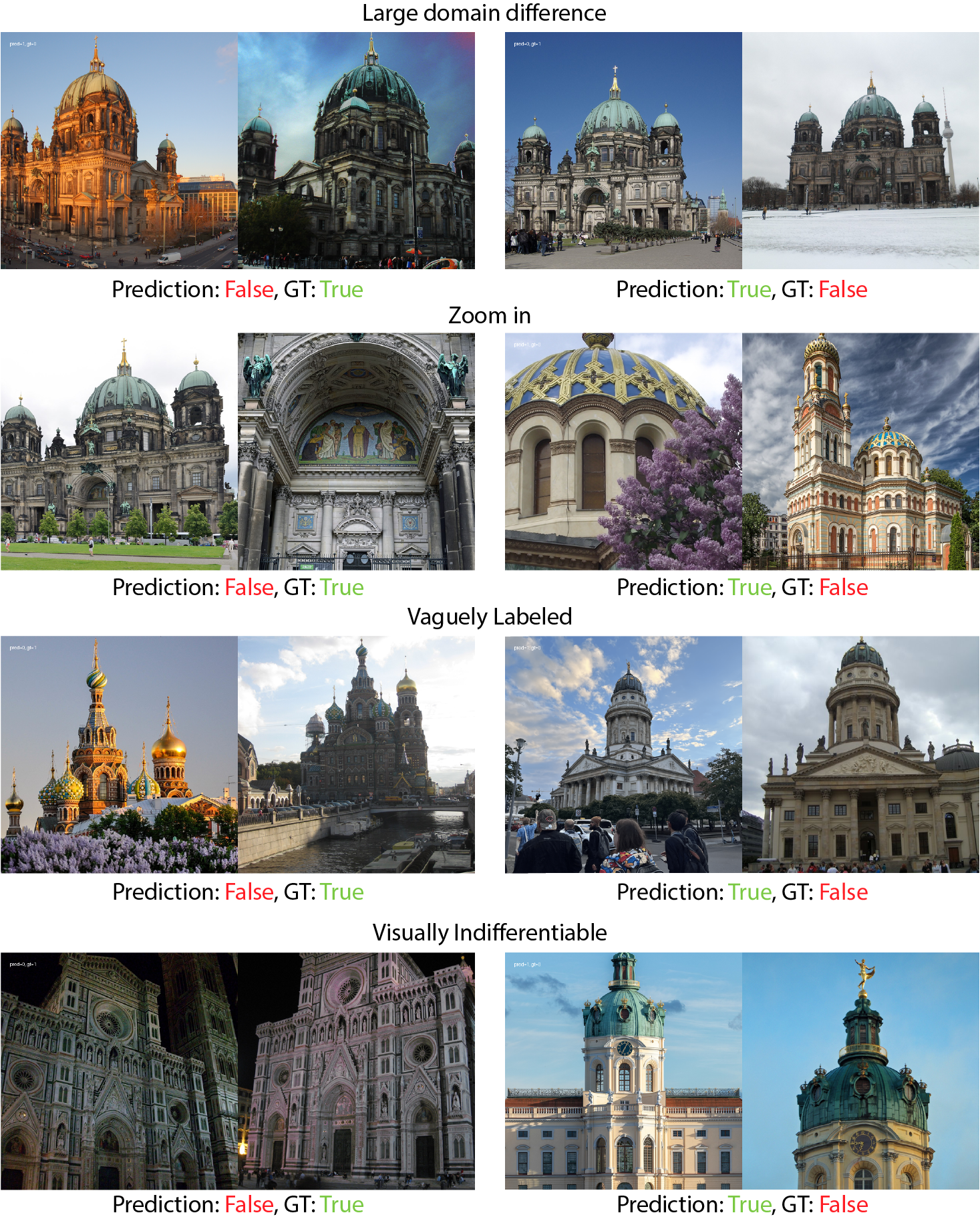}
      \caption{\textbf{Failure cases.}} 
      \label{fig:failure-cases}
\end{figure}